\documentclass{article}
\usepackage{spconf,amsmath,amssymb,graphicx,hyperref}
\usepackage{booktabs}

\title{MOTIONWEAVE: LEARNING MOTION-CENTERED FUTURE DYNAMICS FOR VISION-LANGUAGE-ACTION POLICIES}
\name{Jingqiu Wang, Yan Wang\sthanks{Corresponding author: yanwang@dase.ecnu.edu.cn.}}
\address{School of Data Science and Engineering\\
East China Normal University, Shanghai, China}
\begin{document}
\ninept
\maketitle
\begin{abstract}
Vision-Language-Action (VLA) models have recently incorporated world models to provide richer dynamic supervision beyond sparse action labels. However, explicitly predicting future images or videos may include control-irrelevant appearance, while guidance derived from holistic future visual representations and shared global action features may fail to establish timestep-specific correspondence between actions and local visual changes. To address this issue, we propose MotionWeave, a motion-centric future-dynamics framework for action-chunk prediction with two modules: the Action-Induced Motion Grounder (AIMG) and the Horizon Residual Composer (HRC). Specifically, AIMG conditions on action and proprioceptive representations to construct horizon-specific queries that localize interaction regions associated with each future action timestep from current visual tokens. HRC extracts differences between interaction representations at adjacent horizons, encodes them as temporal motion cues, and injects them into action tokens through a gated residual. During training, robot-arm masks rendered from future frames are used to construct KL-based motion-grounding supervision, while inference uses only the current observation. On six MetaWorld tasks, MotionWeave achieves a 75.3\% average success rate, an absolute gain of 8.6\% over $\pi_0$ (66.7\%), especially on sustained-interaction tasks. Our code is available at \url{https://github.com/autu-mn/MotionWeave}.
\end{abstract}
\section{Introduction}
\label{sec:intro}

Vision-Language-Action (VLA) models unify visual perception, language
understanding, and action generation, providing an effective paradigm for
learning generalizable robot policies~\cite{rt1,rt2,octo,openvla,pi0,openx}. Existing approaches typically build
on pretrained vision-language models~\cite{clip,vit,qwen2_5} and use robot action data to predict
continuous control signals directly from current observations. However,
low-dimensional and sparse action supervision mainly constrains the final
control output, making it difficult to explicitly characterize which visual
regions change during manipulation and how these changes evolve along an
action sequence. Consequently, the visual representations may lack the
local interaction information and temporal cues required for precise
manipulation.

\begin{figure*}[t]
  \centering
  \includegraphics[width=0.88\textwidth]{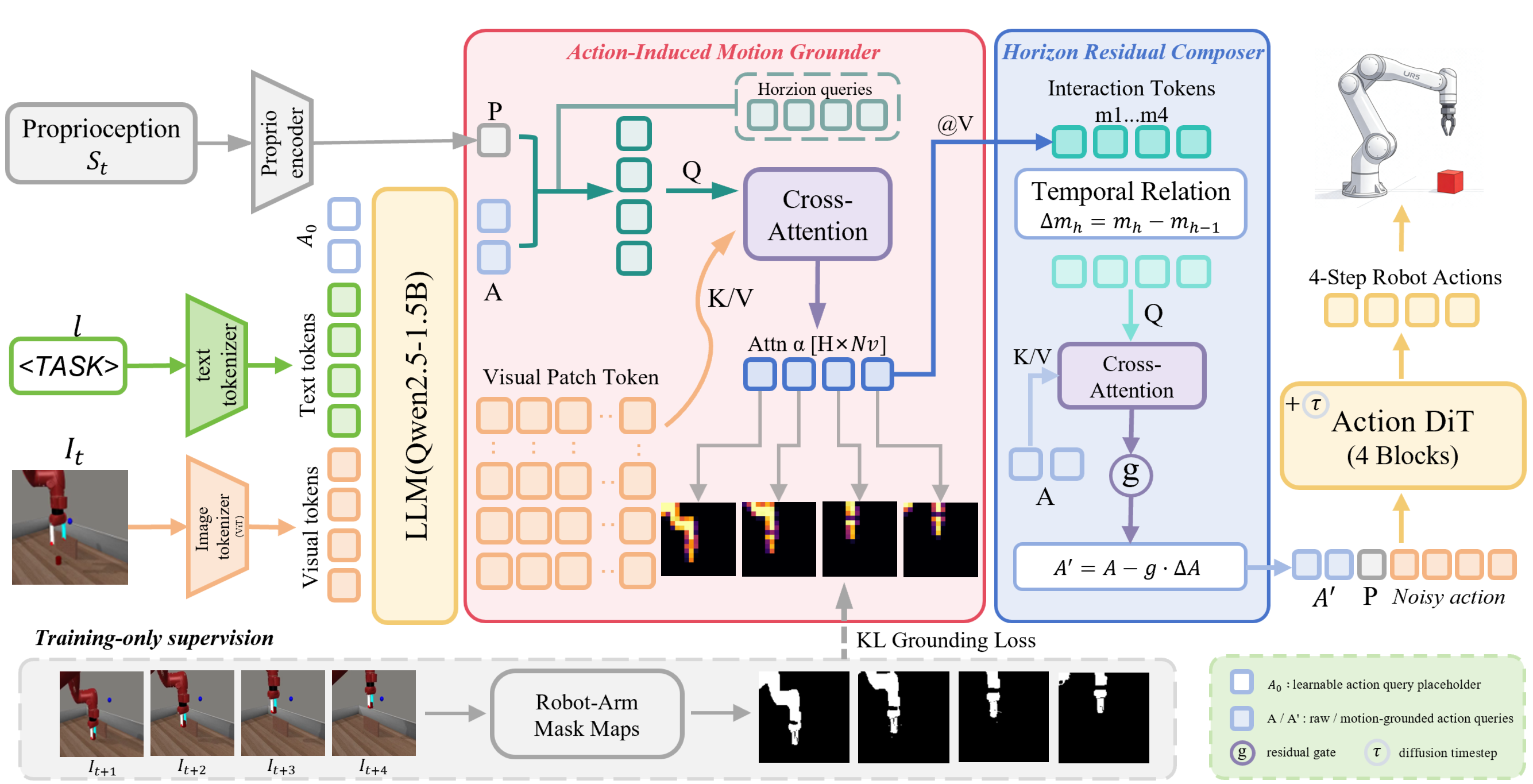}
  \caption{Overview of MotionWeave. The current observation $I_t$ and instruction $l$ are tokenized into visual tokens and text tokens, concatenated with a learnable action query placeholder $A_0$, and jointly processed by LLM to produce action tokens $A$; a separate proprioceptive encoder maps $S_t$ to $P$. The Action-Induced Motion Grounder (AIMG) combines $A$ and $P$ into horizon-specific queries that attend over visual tokens $V$ to produce interaction tokens $M$ with spatial attention $\alpha$. The Horizon Residual Composer (HRC) injects adjacent-horizon differences of $M$ into $A$ via gate $g$ to obtain $A'$, decoded by Action DiT into a 4-step chunk. Robot-arm masks~\cite{robotengine} supervise $\alpha$ via KL loss in training only.}
  \label{fig:architecture}
\end{figure*}

To obtain richer dynamic supervision, recent studies have introduced world models into VLA by predicting future images, videos, or latent representations~\cite{unipi,dreamerv3,worldvla,videopredictionpolicy,dreamvla,wog,fastwam}. WorldVLA~\cite{worldvla} and DreamVLA~\cite{dreamvla} explicitly reconstruct or densely imagine complete future scenes, inevitably entangling control-irrelevant static appearance and background; WoG~\cite{wog} and Fast-WAM~\cite{fastwam} relax pixel-level reconstruction into implicit or test-time-free guidance, yet still supervise the holistic evolution of future visual observations rather than horizon-specific local changes. Meanwhile, action chunking typically decodes a sequence from shared global features~\cite{act,diffusionpolicy}, leaving different timesteps with homogeneous visual evidence and no explicit timing-to-consequence correspondence~\cite{tracevla,atm}. No unified path yet exists from localizing action-relevant regions to step-wise control via their temporal evolution.

Based on these observations, we argue that VLA should not redundantly reconstruct complete future visual observations, but should instead explicitly localize action-relevant dynamic regions and exploit their temporal evolution to guide chunk generation. To this end, we propose \textbf{MotionWeave}, a motion-centered future dynamics learning framework composed of two tightly coupled modules: the \textbf{Action-Induced Motion Grounder (AIMG)} for horizon-specific region localization, and the \textbf{Horizon Residual Composer (HRC)} for fine-grained motion extraction and injection.

AIMG conditions on action representations and proprioceptive states to construct time-indexed queries, attending to current visual tokens to localize regions likely to change at each future horizon and producing interaction representations aligned with the action timeline. HRC then differences adjacent-horizon interaction representations to encode direction and magnitude of change, injecting them into the corresponding action representation via a gated residual. Together they close the loop from region localization, through motion cues, to step-wise control. Future observations provide dynamic supervision only during training and are
not used for inference. The policy therefore retains direct action prediction
at inference time without reconstructing complete future visual observations or incurring
additional inference overhead.

The main contributions of this paper are as follows:
\begin{itemize}
\item We propose the \textbf{Action-Induced Motion Grounder (AIMG)}, which constructs action- and state-conditioned horizon queries and uses future robot-region supervision to shape their attention over the current VLM visual tokens. This provides different action horizons with spatially differentiated visual evidence.

\item We propose the \textbf{Horizon Residual Composer (HRC)}, which represents the temporal evolution of the grounded interaction features through adjacent-horizon differences and injects the resulting cues into action representations through a gated residual update.

\item We achieve state-of-the-art performance on six MetaWorld tasks, achieving an average success rate of 75.3\%, an absolute gain of 8.6\% over $\pi_0$.
\end{itemize}

\section{Methodology}
\label{sec:method}

\subsection{Overall Framework}
\label{ssec:overall}

We consider vision-language-action control with action chunking. At time step $t$, the policy is conditioned on the current RGB observation $I_t$, language instruction $l$, and proprioceptive state $S_t$, and predicts a future action chunk $\hat{a}_{t:t+H-1}$ of length $H=4$, i.e., the conditional distribution $p(\hat{a}_{t:t+H-1}\mid I_t,l,S_t)$. Future RGB observations $I_{t+1:t+H}$ are used only during training to construct horizon-specific motion supervision and are unavailable during inference. For simplicity, we omit the time index $t$ below.

As shown in Fig.~\ref{fig:architecture}, image, text, and proprioception are processed through three branches. InternVL3-2B encodes $I_t$ and $l$ into visual and text tokens, which are concatenated with a learnable action-query placeholder $A_0$ and jointly processed by its Qwen2.5-series language decoder. The model outputs action tokens $A\in\mathbb{R}^{N_a\times D}$ at the $A_0$ positions and spatial visual tokens $V=\{v_n\}_{n=1}^{N_v}$. Here $N_v$ is the number of visual tokens actually output by the InternVL3-2B forward pass (in our implementation $N_v=256$, corresponding to a $16\times 16$ spatial grid); both the AIMG attention $\alpha\in\mathbb{R}^{H\times N_v}$ and the motion-supervision masks are built on this same grid to ensure strict alignment among the three. Separately, a Proprio encoder maps $S_t$ to the proprioceptive embedding $P$; $P$ does not enter the language model and is supplied directly to the subsequent modules. The overall process is

\begin{equation}
(A,V,P)\xrightarrow{\text{AIMG}}(M,\alpha)
\xrightarrow{\text{HRC}}A'
\xrightarrow{\text{Action DiT}}\hat{a}_{t:t+H-1}.
\label{eq:pipeline}
\end{equation}

Here, AIMG retrieves horizon-specific evidence from $V$ to produce $M$ and $\alpha$, HRC writes the temporal evolution of $M$ back into $A$ to obtain $A'$, and Action DiT decodes $\hat{a}$. Both modules are jointly trained with the action and motion-grounding objectives.

\subsection{Action-Induced Motion Grounder (AIMG)}
\label{ssec:grounder}

AIMG does not reconstruct future RGB frames. Instead, it identifies which of the current $N_v$ visual-token positions will change within the next $H$ steps. Let $e_h$ be the learnable horizon embedding for the $h$-th future timestep, where $h=1,\ldots,H$. After LayerNorm and linear projection, the action tokens $A$ and proprioceptive embedding $P$ are mapped into a common motion space of dimension $d_m$, and horizon-specific queries are formed as

\begin{equation}
\begin{aligned}
q_h = e_h + \operatorname{Mean}(\operatorname{Proj}_a(A)) + \operatorname{Proj}_p(P),
\end{aligned}
\label{eq:query}
\end{equation}
where $\operatorname{Mean}(\cdot)$ is taken over the projected action queries so that each horizon query carries global action guidance. A lightweight residual transformation then refines $q_h$. Let $\operatorname{Proj}_v$ denote the shared visual-value projection. The attention and interaction representations~\cite{vaswani2017attention} are:

\begin{equation}
\alpha_{h,n} = \operatorname{softmax}_{n}\!\left(
\frac{q_h^{\top}\operatorname{Proj}_v(v_n)}{\sqrt{d_m}}\right),
\label{eq:attention}
\end{equation}
\begin{equation}
m_h = \operatorname{LN}\!\left(q_h +
\sum_{n=1}^{N_v}\alpha_{h,n}\operatorname{Proj}_v(v_n)\right).
\label{eq:interaction}
\end{equation}

The spatial attentions form $\mathrm{Attn}\in\mathbb{R}^{H\times N_v}$ and receive KL motion-grounding supervision. The aggregated interaction tokens $M=[m_1,\ldots,m_H]$ are passed to HRC. All $m_h$ come from the current observation and are differentiated by $e_h$, aligning them with future action timesteps.

During training, each sample provides $I_t,I_{t+1},\ldots,I_{t+H}$. For each $h$, we render the corresponding robot-arm mask with Robot Engine~\cite{robotengine} and adopt the mask of $I_{t+h}$ as the horizon-specific motion target: it marks the arm region executing the $h$-th future action while suppressing static background, texture, and other task-irrelevant appearance, so the mask itself serves as the motion of interest. Each binary mask is downsampled to the $16\times 16$ visual-token grid, smoothed with a small constant for numerical stability, and normalized to yield the target distribution $\mu_h$. This construction requires no manual annotation and reuses existing expert trajectories with off-the-shelf mask generation.

\subsection{Horizon Residual Composer (HRC)}
\label{ssec:composer}

The interaction representations produced by AIMG are independent and do not explicitly encode scene evolution. After projecting $M$ into the HRC space, we define

\begin{equation}
\Delta m_h=m_h-m_{h-1},\qquad m_0:=m_1,
\label{eq:delta}
\end{equation}
so that $\Delta m_1=0$ and later differences encode the direction and magnitude of change relative to the preceding timestep. HRC uses motion differences as queries and action queries as keys and values in cross-attention, allowing each local-evolution cue to probe the action query that needs correction:

\begin{equation}
U=\operatorname{MHA}\left(\operatorname{Proj}_m(\Delta M),
\operatorname{Proj}_a(A),\operatorname{Proj}_a(A)\right),
\label{eq:composer_attn}
\end{equation}
where $\Delta M=[\Delta m_1,\ldots,\Delta m_H]$ and $\operatorname{Proj}_a$ is the HRC-side action projection, independent of the AIMG-side projection~\cite{vaswani2017attention}. The update and proprioception-conditioned residual gate are

\begin{equation}
\Delta A=\operatorname{Proj}_o(U),
\qquad
g=\sigma\!\left(\operatorname{Proj}_g(P)\right),
\label{eq:gate}
\end{equation}
and the motion-grounded action query is obtained by subtractive residual update:

\begin{equation}
A'=A-g\odot\Delta A.
\label{eq:residual}
\end{equation}

Unlike direct overwriting, this form lets each action query absorb only the motion increment associated with its timestep. The correction direction comes from cross-attention over temporal differences, while the gate adaptively controls injection strength and preserves the original control prior in $A$.

\subsection{Training Objective}
\label{ssec:objective}

Conditioned on $A'$, $P$, noisy action chunk $a^{\tau}$, and diffusion timestep $\tau$, Action DiT~\cite{dit} predicts $\hat{a}_{t:t+H-1}$ with the flow-matching~\cite{flowmatching} loss $\mathcal{L}_{\mathrm{act}}$. AIMG aligns each $\alpha_h$ with its mask distribution $\mu_h$:

\begin{equation}
\mathcal{L}_{\mathrm{motion}}=\frac{1}{H\log N_v}
\sum_{h=1}^{H}\mathrm{KL}\!\left(\mu_h\,\|\,\alpha_h\right),
\label{eq:motion_loss}
\end{equation}
where $\mathrm{KL}(\mu_h\,\|\,\alpha_h)$ penalizes attention mass placed outside the arm region at horizon $h$, and division by $\log N_v$ normalizes the loss by the entropy of the uniform distribution over visual-token positions. The complete objective is

\begin{equation}
\mathcal{L}_{\mathrm{total}}
=\mathcal{L}_{\mathrm{act}}
+\lambda_{\mathrm{motion}}\mathcal{L}_{\mathrm{motion}},
\label{eq:total_loss}
\end{equation}
with $\lambda_{\mathrm{motion}}=0.05$ in the implemented configuration. At inference time, the policy receives only $I_t$, $l$, and $S_t$: the mask-rendering and KL branches are removed, and DiT denoising outputs four actions without reconstructing future visual observations.

\begin{figure*}[t]
  \centering
  \includegraphics[width=0.92\textwidth]{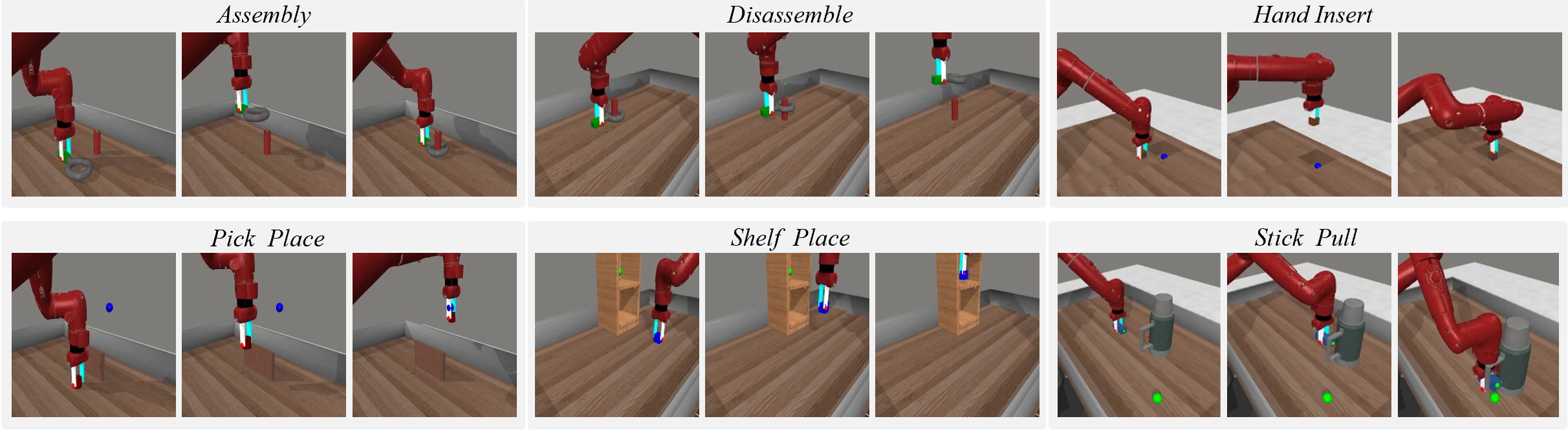}
  \caption{Qualitative rollouts on six tasks. Each row shows key frames from observation to successful completion.}
  \label{fig:qualitative}
\end{figure*}
\begin{table*}[t]
  \centering
  \caption{Success rates (\%) on six MetaWorld tasks. All methods share the evaluation random seeds, and bold denotes the best result for each task. The ``Source'' column reports the publication status of each baseline.}
  \label{tab:main}
  \footnotesize
  \setlength{\tabcolsep}{4pt}
  \begin{tabular*}{\textwidth}{@{\extracolsep{\fill}}lcccccccc}
    \toprule
    Method & Source & Pick Place & Disassemble & Stick Pull & Assembly & Shelf Place & Hand Insert & Avg. \\
    \midrule
    $\pi_0$~\cite{pi0} & RSS'25 & \textbf{72.0} & 52.0 & 68.0 & 76.0 & 64.0 & \textbf{68.0} & 66.7 \\
    DreamVLA~\cite{dreamvla} & NeurIPS'25 & 60.0 & 68.0 & 72.0 & 70.0 & 56.0 & 64.0 & 65.0 \\
    WoG~\cite{wog} & ICML'26 & 28.0 & 60.0 & 68.0 & 24.0 & 64.0 & 60.0 & 50.7 \\
    Fast-WAM~\cite{fastwam} & arXiv'26 & 44.0 & 56.0 & 16.0 & 64.0 & 56.0 & 60.0 & 49.3 \\
    MotionWeave (ours) & --- & 60.0 & \textbf{92.0} & \textbf{76.0} & \textbf{80.0} & \textbf{76.0} & \textbf{68.0} & \textbf{75.3} \\
    \bottomrule
  \end{tabular*}
\end{table*}

\section{Experiments}
\label{sec:exp}

\subsection{Experimental Setup}
\label{ssec:expsetup}

\textbf{Tasks and data}. We select six challenging MetaWorld~\cite{metaworld} tasks: Pick Place, Disassemble, Stick Pull, Assembly, Shelf Place, and Hand Insert, covering grasping, assembly, pulling, insertion, and long-horizon placement. Each task contains 25 expert trajectories of 175 timesteps, for 150 trajectories and 26{,}250 frames in total. The input consists of a single-view $224 \times 224$ RGB image, a language instruction, and the proprioceptive state, with supervision on the next 4-step actions ($H=4$).

\textbf{Model and training}. Our method MotionWeave builds on InternVL3-2B, whose language side uses a Qwen2.5-series decoder~\cite{internvl3,qwen2_5}, with full fine-tuning. Training runs for 20k steps on two 48GB NVIDIA A40 GPUs with FSDP full sharding and gradient checkpointing; the per-GPU batch size is 8 and the global batch size is 16. The optimizer uses learning rate $1 \times 10^{-5}$, weight decay 0, and gradient clipping 1.0. The action head is a 12-layer, 768-dimensional, 12-head DiT~\cite{dit} with a 4-step action horizon, a flow-matching objective~\cite{flowmatching}, 100 training diffusion steps, and 10 inference steps. AIMG uses 4 horizon queries with hidden dimension 512, supervised by future motion maps from $t+1$ to $t+4$; HRC uses 768-dimensional, 8-head attention to update action tokens through a gated residual. The total loss is $\mathcal{L}_{\mathrm{act}}+0.05\,\mathcal{L}_{\mathrm{motion}}$; future frames are used only during training.

\textbf{Evaluation protocol}. We evaluate each task over 25 episodes using initialization seeds shared across methods, with a 175-step limit and replanning every 2 executed steps. Success follows the official MetaWorld binary criterion; we report the macro-average success rate across six tasks.

\subsection{Main Results}
\label{ssec:mainres}

Table~\ref{tab:main} compares MotionWeave with $\pi_0$~\cite{pi0}, DreamVLA~\cite{dreamvla}, WoG~\cite{wog}, and Fast-WAM~\cite{fastwam} under the shared protocol. The baselines cover the two alternatives discussed in Sec.~\ref{sec:intro}: direct action mapping from global features ($\pi_0$), and action guidance based on modeling complete future visual observations (DreamVLA, WoG, Fast-WAM). Predicting complete future visual observations introduces abundant static background, texture, and appearance content that is redundant for action decisions, whereas MotionWeave directly models action-relevant motion regions, directions, and magnitudes, concentrating limited capacity on how the robot and objects will change. MotionWeave reaches an average success rate of 75.3\%, yielding an absolute gain of 8.6\% over the strongest baseline $\pi_0$, and leads on five of six tasks. The advantage is largest on sustained-interaction tasks, including Disassemble (92.0\% vs.\ 68.0\%) and Shelf Place (76.0\% vs.\ 64.0\%). On Pick Place, $\pi_0$ performs better (72.0\% vs.\ 60.0\%), where the target displacement is short and global features already suffice. Fig.~\ref{fig:qualitative} shows rollouts on all six tasks from observation to successful completion.

\subsection{Ablation Studies}
\label{ssec:ablation}

Table~\ref{tab:ablation_components} separates the effects of horizon-specific grounding, explicit attention supervision, and temporal composition. Adding the AIMG architecture without the grounding objective improves the average success rate from 58.0\% to 62.0\%. This modest gain suggests that horizon embeddings and action-conditioned queries already introduce useful temporal differentiation, but the action loss alone provides only indirect spatial guidance. Supervising the horizon-wise attention distributions further raises the result to 70.0\%, supporting the central role of AIMG: future robot-region targets shape how each horizon query retrieves control-relevant evidence from the current VLM visual tokens. Finally, HRC improves the result from 70.0\% to 75.3\%. Thus, horizon-specific localization is useful on its own, while explicitly composing differences between adjacent grounded representations provides additional temporal information for action generation. Together, the ablation follows the intended design path from spatial grounding to temporal differencing and action conditioning.

\begin{table}[htbp]
  \centering
  \caption{Component ablation results for MotionWeave. We report macro-average success (\%) over the six tasks. Each row adds one component relative to the preceding configuration.}
  \label{tab:ablation_components}
  \footnotesize
  \renewcommand{\arraystretch}{1.25}
  \setlength{\tabcolsep}{10pt}
  \begin{tabular}{l @{\hspace{2.5cm}} c}
    \toprule
    Variant & Avg. \\
    \midrule
    Baseline & 58.0 \\
    + AIMG w/o $\mathcal{L}_{\mathrm{mot}}$ & 62.0 \\
    + AIMG & 70.0 \\
    + AIMG + HRC & 75.3 \\
    \bottomrule
  \end{tabular}
\end{table}

\begin{figure}[htbp]
  \centering
  \includegraphics[width=\columnwidth]{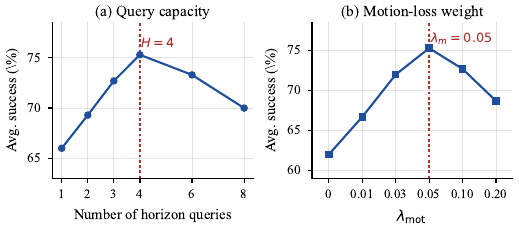}
  \caption{Hyperparameter curves in average success (\%). (a) Horizon query count peaks at 4 with 75.3\%, aligned with the action chunk length $H=4$, falling back to 73.3\%/70.0\% at 6/8 queries. (b) Motion loss weight $\lambda_{\mathrm{motion}}$ peaks at 0.05 with 75.3\%, falling back to 72.7\%/68.7\% at 0.10/0.20.}
  \label{fig:ablation}
\end{figure}

Fig.~\ref{fig:ablation} reports sensitivity to two hyperparameters: the horizon query count peaks at 4 with 75.3\%, aligned with the action chunk length $H=4$, so one query per action step provides a clearer horizon-action correspondence; redundant queries introduce tokens with competing attention. The motion loss weight peaks at $\lambda_{\mathrm{motion}}=0.05$, after which success decreases.

\section{Conclusion}
\label{sec:conclusion}

MotionWeave enables action-chunk policies to represent the visual changes relevant to each action step without reconstructing complete future visual observations. AIMG uses action- and proprioception-conditioned horizon queries to localize interaction regions in current visual tokens, while HRC encodes changes between adjacent horizons and injects these motion cues into the corresponding action tokens through a gated residual. Future-frame masks supervise horizon-specific grounding during training, allowing the policy to predict actions directly from the current observation at inference. Ablation results confirm that both horizon-wise grounding and temporal differencing contribute to the overall performance gain. On six MetaWorld tasks, MotionWeave achieves an average success rate of 75.3\%, yielding an absolute gain of 8.6\% over the strongest baseline $\pi_0$, and improves performance on five tasks. These results show that aligning localized motion evidence with action horizons improves visuomotor control, especially in tasks that require sustained interaction. Future work will investigate whether the motion-grounded representations learned in simulation can generalize to real-robot settings with different viewpoints, object appearances, contact dynamics, and execution noise. Extending the evaluation to unseen tasks and hardware platforms will further clarify the transferability of horizon-specific motion guidance.

\clearpage
\bibliographystyle{IEEEbib}
\bibliography{strings,refs}

\end{document}